\documentclass[letterpaper]{article} 
\usepackage{aaai2026}  
\usepackage{times}  
\usepackage{helvet}  
\usepackage{courier}  
\usepackage[hyphens]{url}  
\usepackage{graphicx} 
\usepackage{natbib}  
\usepackage{caption} 
\usepackage{algorithm}
\usepackage{algpseudocode}
\usepackage{enumitem}
\usepackage{amsmath}
\usepackage{booktabs}
\usepackage{amssymb} 

\usepackage{xcolor}
 
\usepackage{newfloat}
\usepackage{listings}
\DeclareCaptionStyle{ruled}{labelfont=normalfont,labelsep=colon,strut=off} 
\floatstyle{ruled}
\newfloat{listing}{tb}{lst}{}
\floatname{listing}{Listing}
\nocopyright

\title{Generative Modeling with Multi-Instance Reward Learning for E-commerce Creative Optimization}
\author{
    Qiaolei Gu\textsuperscript{\rm 1}\equalcontrib,
    Yu Li\textsuperscript{\rm 2,3}\equalcontrib,
    DingYi Zeng\textsuperscript{\rm 1},
    Lu Wang\textsuperscript{\rm 1}\thanks{Corresponding author: Lu Wang},
    Ming Pang\textsuperscript{\rm 1},
    Changping Peng\textsuperscript{\rm 1},
    Zhangang Lin\textsuperscript{\rm 1},
    Ching Law\textsuperscript{\rm 1},
    Jingping Shao\textsuperscript{\rm 1}
}
\affiliations{
    \textsuperscript{\rm 1}JD.com\\
    \textsuperscript{\rm 2}Institute of Computing Technology, Chinese Academy of Sciences\\
    \textsuperscript{\rm 3}University of Chinese Academy of Sciences\\
    guqiaolei1@jd.com,
    liyu@ict.ac.cn,
    \{zengdingyi1,wanglu241,pangming8,pengchangping,linzhangang,lawching,shaojingping\}@jd.com

}

\usepackage{bibentry}

\begin{document}

\maketitle

\begin{abstract}

In e-commerce advertising, selecting the most compelling combination of creative elements --- such as titles, images, and highlights --- is critical for capturing user attention and driving conversions. However, existing methods often evaluate creative components individually, failing to navigate the exponentially large search space of possible combinations. To address this challenge, we propose a novel framework named GenCO that integrates generative modeling with multi-instance reward learning. Our unified two-stage architecture first employs a generative model to efficiently produce a diverse set of creative combinations. This generative process is optimized with reinforcement learning, enabling the model to effectively explore and refine its selections. Next, to overcome the challenge of sparse user feedback, a multi-instance learning model attributes combination-level rewards, such as clicks, to the individual creative elements. This allows the reward model to provide a more accurate feedback signal, which in turn guides the generative model toward creating more effective combinations. Deployed on a leading e-commerce platform, our approach has significantly increased advertising revenue, demonstrating its practical value. Additionally, we are releasing a large-scale industrial dataset to facilitate further research in this important domain.

\end{abstract}

\section{Introduction}

In the competitive landscape of e-commerce, the effectiveness of an advertisement is determined not only by the product it promotes but also by the creative with which it is presented~\citep{chen2021efficient}.
High-quality ad creatives are a key differentiator, significantly enhancing a product's appeal and driving user engagement. 
As a result, creative optimization has become a critical focus area for modern e-commerce platforms, complementing established strategies like click-through rate (CTR) prediction to boost advertising revenue and improve the user experience. 
With recent advances in AI-generated content (AIGC), the ability to produce a vast and diverse array of creative materials has become more accessible, further highlighting the importance of systematically optimizing these assets~\citep{wei-etal-2022-creater,Jiang2023AdSEEIT}.

\begin{figure}[t]
    \centering
    \includegraphics[width=\linewidth]{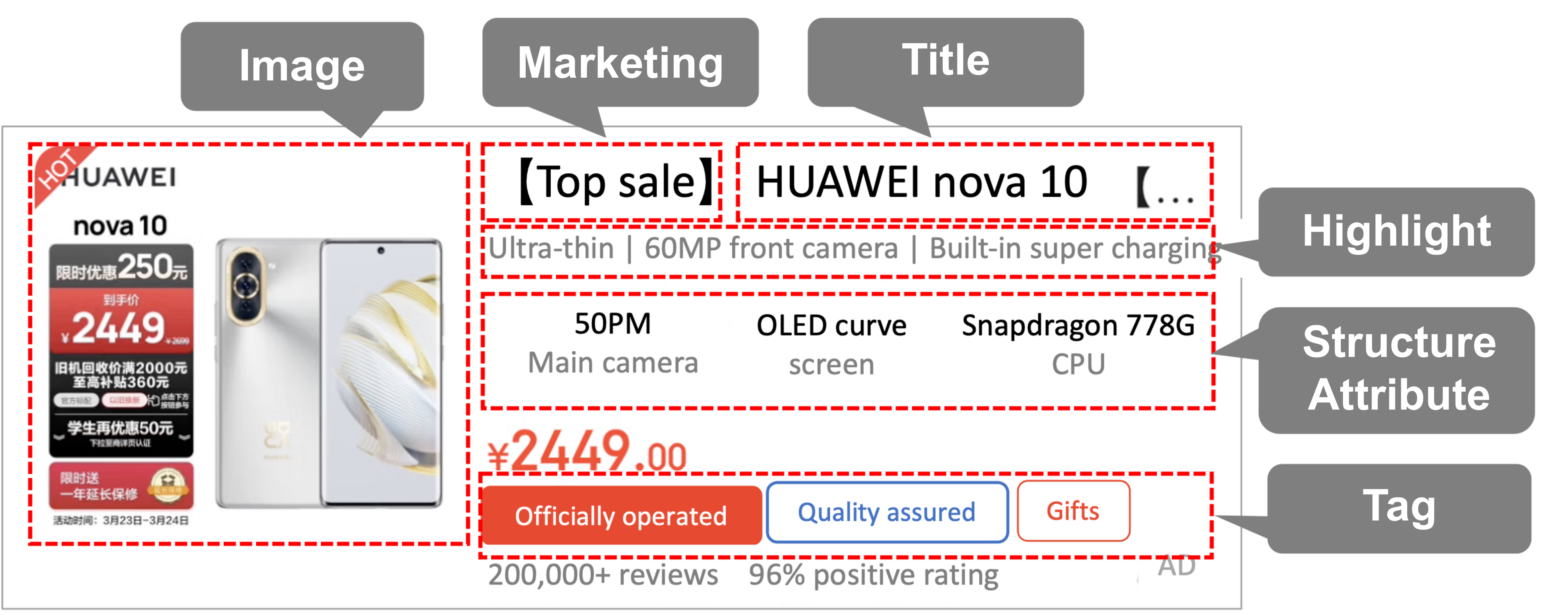}
    \caption{Illustration of a product card. Red dashline boxes indicate different components. For each component, there are multiple candidate elements to be selected.
    }
    \label{fig:product_card}
\end{figure}

An e-commerce ad creative is a composite of multiple building blocks, which we define as components.
These components represent distinct sections of the ad, such as the Title, Image, Highlights, and descriptive Tags.
For each component, the e-commerce platform can choose from a pool of candidate options, which we term elements.
For instance, the Image component might have several different pictures available for the same product.
As illustrated in Figure~\ref{fig:product_card}, a complete ad shown to a user is therefore a specific combination of these elements, formed by selecting one or several elements for each component.
The central issue is not merely to create good individual elements, but to identify the most compelling combination of elements that maximizes user engagement and, ultimately, conversion rates.

To address this issue, there involves three significant challenges.
First is the combinatorial explosion: as the number of elements for each component increases, the total number of unique ad combinations grows exponentially, rendering exhaustive evaluation computationally infeasible in a real-time system.
To manage this complexity, previous work has often circumvented the problem by focusing only on a single component,
such as image selection \citep{wang2021hybrid,you2023image},
or by using pre-determined pairs like image-title combinations \citep{lin2022joint},
which fails to explore the full creative combination space.
While several works consider the selection of multiple components~\citep{yang2024parallel,chen2103automated,chen2021efficient},
their approach relies on a brute-force enumeration of all possible combinations.
Such a method is only feasible due to their use of small candidate pools and is intractable in real-world scenarios with extensive creative element candidates,
thus exacerbating the challenge of combinatorial explosion.

Second, the challenge is compounded by sparse user feedback.
The sheer volume of possible combinations, coupled with the dynamic nature of creative elements---which are frequently updated for events like seasonal promotions---means that any single combination receives very few impressions.
This data scarcity makes it difficult to reliably assess creative effectiveness~\citep{zhao2019you}.
In response, many studies have framed creative selection as an exploration--exploitation problem,
seeking to balance the discovery of novel creatives with the utilization of proven ones~\citep{chen2103automated,wang2021hybrid,chen2021efficient}.
Although valuable, those work has largely neglected the potential of advanced model architectures and holistic system design to effectively address data sparsity in practical applications.

Third, a crucial challenge lies in disentangling creative impact from product itself.
An ad's performance is driven by both the product's intrinsic desirability and the effectiveness of its creative presentation.
By isolating the influence of the creative, we can achieve a more precise and nuanced evaluation of different creative elements, which is the key to selecting the optimal combination. However, this is non-trivial.
A significant limitation of prior research is its tendency to focus exclusively on the creatives themselves, neglecting the critical interaction between creatives and CTR prediction models~\citep{wei2022towards,koren2020dynamic,mishra2020learning}.
In any practical advertising system, a creative's effectiveness is inseparable from the CTR model's predictions; thus, optimizing them in concert is essential for achieving maximum performance.

To address these challenges, we propose GenCO, a novel \textbf{Generative}  framework that combines generative modeling and multi-instance reward learning for \textbf{Creative Optimization}. 
Our unified two-stage architecture is specifically designed to efficiently explore the exponentially large combinatorial space of creative combinations while effectively learning from sparse user feedback. In the first stage, a generative model is employed to produce a diverse set of promising creative combinations, with reinforcement learning guiding the selection process to disentangle the impact of creative elements from inherent product appeal by  optimizing reward signals. In the second stage, a multi-instance learning model attributes combination-level rewards to individual creative elements, enabling more accurate feedback and further refining the generative process. 
When deployed on a leading e-commerce platform, GenCO has demonstrated substantial improvements in advertising revenue.

Our main contributions are summarized as follows:

\begin{itemize}
    \item We propose GenCO, a novel generative framework that systematically addresses the combinatorial explosion in creative selection for e-commerce advertising. Our approach leverages a non-autoregressive, context-aware generative model to efficiently explore and generate diverse, high-potential creative combinations while modeling complex dependencies among creative elements.
    \item To tackle the challenges of data sparsity and disentangling creative effectiveness from product appeal, we integrate reinforcement learning for end-to-end reward optimization and multi-instance learning to attribute combination-level user feedback to individual creative elements, enabling more robust  learning.
    \item Our unified two-stage architecture decouples candidate combination generation from their evaluation, ensuring both scalability and low-latency inference suitable for deployment in real-world, large-scale commercial systems.
    \item To facilitate further research in creative optimization, we release a large-scale industrial dataset containing candidate sets for multiple creative components, the first of its kind in this domain.
\end{itemize}

\begin{figure*}[t]
    \centering
    \includegraphics[width=1\linewidth]{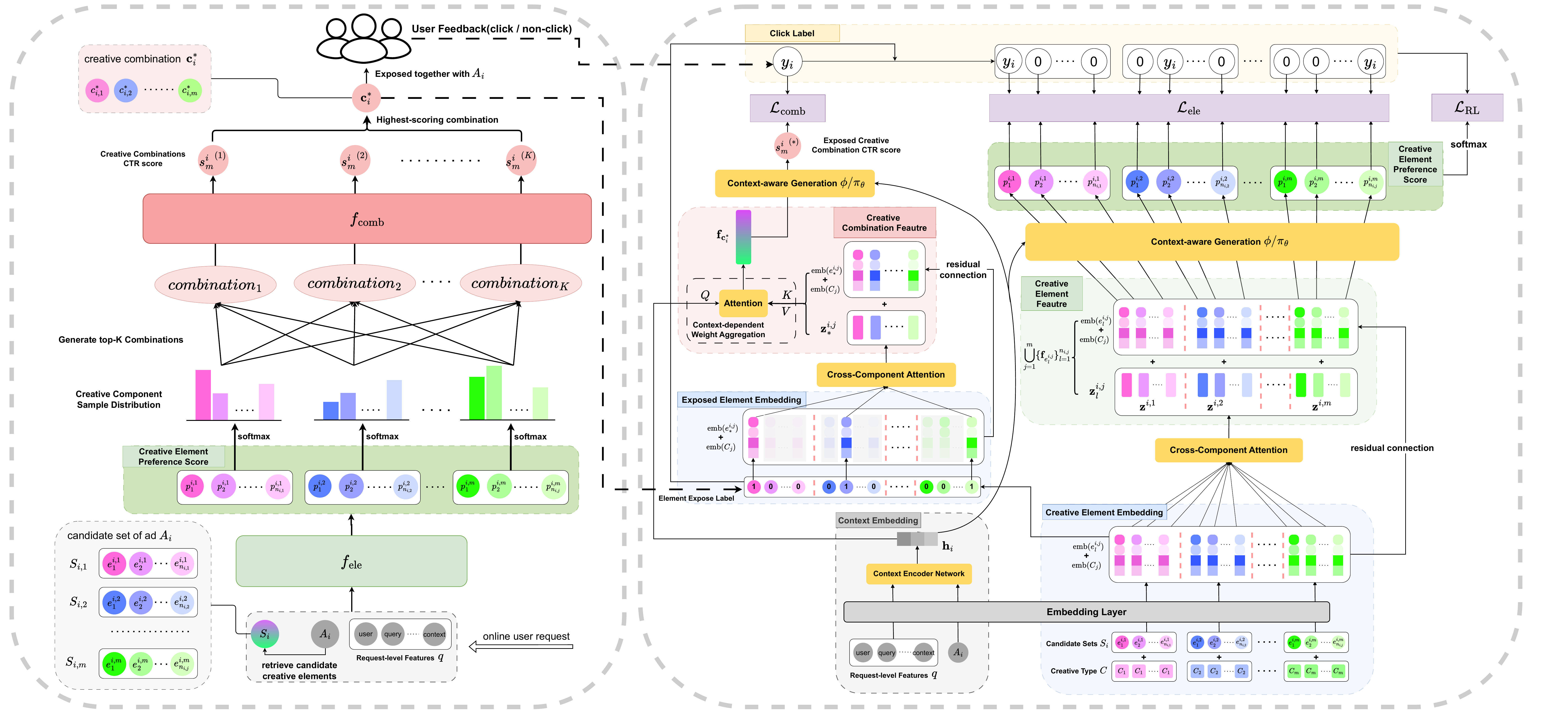}
    \caption{
    Current system framework of GenCO. Left hand side is online serving of GenCO. Right hand side is offline network modeling and training details. User feedback (click or non-click) is used to optimize both context-aware generation function $f_{\mathrm{ele}}$ and combination-level reward function $f_{\mathrm{comb}}$.
    }
    \label{fig:mainframework}
\end{figure*}

\section{Related Works}

Initial approaches to creative selection framed the task as a combinatorial multi-armed bandit problem, which involves selecting an optimal subset of arms simultaneously~\citep{cesa2012combinatorial}. This formulation is analogous to problems such as web page selection for display~\citep{chen2013combinatorial} and other top-k selection challenges~\citep{rejwan2020top}. Consequently, numerous bandit algorithms with structured arms found practical application in online advertising and personalized recommendation systems~\citep{li2010exploitation,chapelle2011empirical}. More recently, the focus has shifted toward treating creative optimization as a click-through rate (CTR) prediction problem, a perspective that has attracted significant research interest~\citep{chen2016deep,liu2020category}.
Much of the existing literature on creative selection concentrates on optimizing individual components, like images or headlines, in isolation~\citep{yang2024new,wang2021hybrid,you2023image}. While this line of work has produced valuable insights into element-level effectiveness, it often fails to address the combinatorial explosion that occurs when assembling individual components into complete advertisements~\citep{zhao2019you,lin2022joint,mita2023striking}.
Furthermore, a significant body of research has been dedicated to managing the exploration-exploitation trade-off. These studies aim to balance the discovery of novel, high-potential creatives with the utilization of those already known to be effective~\citep{chen2103automated,wang2021hybrid,chen2021efficient}. However, a common limitation in prior work is its narrow focus on the creative assets alone, often neglecting the critical, synergistic interaction between the creatives and the underlying CTR prediction models that determine their delivery~\citep{wei2022towards,koren2020dynamic,mishra2020learning}.

\section{Methodology}

\subsection{Problem Formulation}
The task of creative selection involves determining the optimal combination of creative elements for each candidate advertisement, with the objective of maximizing the expected utility or performance of the ad. After identifying these optimal creative combinations, the results are transmitted to the post-link ranking module. This module is generally designed to maximize overall social welfare by ranking the advertisements, each already associated with its optimized creative set, before presenting them to users.

For each advertisement \( A_i \) and each creative component \( C_j \) --- a creative component represents a creative type, e.g., Image or Title ---
there exists a set of candidate creative elements to choose from.  
This set is denoted as:  
\begin{align}  
S_{i,j} = \{ e_1^{i,j}, e_2^{i,j}, \dots, e_{n_{i,j}}^{i,j} \},
\end{align}  
where \( e_k^{i,j} \) represents the \( k \)-th candidate for creative component \( C_j \) in advertisement \( A_i \),  
and \( n_{i,j} \) is the total number of candidates available for that specific creative component in that advertisement.  

A creative combination for an advertisement \( A_i \) is formed by selecting one candidate from each creative component.  
It is represented as a tuple:  
\begin{align}  
\mathbf{c}_i = (c_{i,1}, c_{i,2}, \dots, c_{i,m}),  
\end{align}  
where \( c_{i,j} \) is a selected candidate from the set \( S_{i,j} \) for each creative component \( j = 1, 2, \dots, m \).  

The set of all possible creative combinations for advertisement \( A_i \) is given by the Cartesian product:  
\begin{align}  
\mathbf{C}_i = S_{i,1} \times S_{i,2} \times \dots \times S_{i,m}.  
\end{align}  
The total number of possible combinations is calculated as:  
\begin{align}  
|\mathbf{C}_i| = n_{i,1} \times n_{i,2} \times \dots \times n_{i,m}.  
\end{align}  

Let \( q \) denote the request.  
In our e-commerce ad setting,  
the effectiveness of the tuple \( (A_i, \mathbf{c}_i, q) \) is measured by the click-through rate (CTR).  
Formally, the creative selection task is to solve the following problem:  
\begin{align}  
\max_{\mathbf{c}_i \in \mathbf{C}_i} \, \operatorname{CTR}(A_i, \mathbf{c}_i, q).  
\end{align}

\subsection{General Architecture Design}
The core challenge in creative element selection lies in the exponential growth of the combinatorial space: for each advertisement $A_i$, and for each creative component $C_j$, there exists a candidate set $S_{i,j}$, and the total number of possible creative combinations grows rapidly with the number and size of these candidate pools. Exhaustively evaluating all combinations is computationally infeasible, especially under real-time constraints.

To address this, we propose a unified two-stage architecture that balances efficiency and effectiveness in both training and inference (see left part of Figure \ref{fig:mainframework}). The first stage, generative modeling, efficiently produces a diverse set of candidate creative combinations by casting element selection as a context-aware generation process. The second stage, reward modeling, evaluates and selects the optimal combination based on predicted click-through rates (CTR).

In the generative modeling stage, given an input tuple
\begin{align}
    (A_i, \{S_{i,j}\}_{j=1}^m, q),
\end{align}
where $A_i$ denotes the advertisement, $\{S_{i,j}\}_{j=1}^m$ are the candidate sets for each of the $m$ creative component, and $q$ encapsulates request-level (e.g., user, query, context) features, the model computes a preference score for each candidate element. The total number of candidate scores is
\begin{align}
    N_{\text{total}} = \sum_{j=1}^m n_{i,j},
\end{align}
where $n_{i,j} = |S_{i,j}|$ is the number of candidates for component $C_j$ in $A_i$. The scores within each creative component are normalized via a softmax function, forming a categorical distribution for efficient, parallel sampling of candidate combinations.

For each element $e_k^{i,j} \in S_{i,j}$, the score is given by
\begin{align}
    p_k^{i,j} = f_{\mathrm{ele}}(q, A_i, S_{i,j}, e_k^{i,j})
\end{align}
where $f_{\mathrm{ele}}$ is a context-aware   generation function.
For combination generation, one candidate per creative component is sampled according to the softmax-normalized scores

yielding a combination $\mathbf{c}_{i} = (c_{i,1}, ..., c_{i,m})$.
Here, for simplicity, we sample one candidate per creative component.
This design enables efficient parallel combination generation.
Our framework, however, can be extended to more complex sampling strategies such as sampling multiple candidates per component or hierarchical generation.

In the reward modeling stage, each generated combination $\mathbf{c}_{i}$ is evaluated by a combination-level reward function  $f_{\mathrm{comb}}$.
For a given advertisement $A_i$, request-level features $q$, and candidate combination $\mathbf{c}_{i}$,
the predicted CTR score is
\begin{align}
    s_m^i = f_{\mathrm{comb}}(q, A_i, \mathbf{c}_{i}).
\end{align}
The combination with the highest predicted CTR is selected as the optimal output:
\begin{align}
    \mathbf{c}_i^* = \arg\max_{\mathbf{c}_{i}}(s_m^i)= \arg\max_{\mathbf{c}_{i}} f_{\mathrm{comb}}(q, A_i, \mathbf{c}_{i}).
\end{align}

Both modules are trained using observed click feedback. For the reward model, exposed pairs $(A_i, \mathbf{c}_{i})$ with click labels $y_i \in \{0, 1\}$ are used to minimize the cross-entropy loss:
\begin{align}
    \mathcal{L}_{\mathrm{comb}} = -\sum_i y_i \log (\mathrm{sigmoid}(s_m^i)).
\end{align}
Similarly, the element-level model $f_{\mathrm{ele}}$ is trained with exposed elements and the same click labels:
\begin{align}
    \mathcal{L}_{\mathrm{ele}} = -\sum_{i,j} y_i \log(\mathrm{sigmoid}(p_k^{i,j})).
\end{align}

This unified two-stage framework decouples the efficient, context-aware generation of diverse candidate combinations from their accurate evaluation, enabling scalable and effective creative selection under strict latency constraints.

\subsection{Generative Element Modeling and Multi-Instance Reward Learning}

The traditional element selection paradigm in creative optimization typically relies on discriminative scoring or ranking, where each candidate element is independently assigned a score according to static or shallow contextual features. However, this approach suffers from two intrinsic limitations: (1) it cannot accurately capture the complex dependencies and contextual interactions among creative elements, leading to suboptimal combination estimation; (2) its reliance on observed feedback for only a minute fraction of possible combinations induces severe data sparsity, resulting in weak supervision and poor generalization, especially in cold-start scenarios.

To address these challenges, we propose a fundamental paradigm shift by reformulating creative element and combination selection as a context-aware generative process. In this framework, the element selection model $f_{\mathrm{ele}}$ is cast as a generative model over the combinatorial action space, analogous to token selection in vocabulary sampling for natural language generation. This transition enables the model to dynamically adjust its action space and state representation based on global and local context, thereby capturing intricate inter-element dependencies. Further details are illustrated in the right part of Figure \ref{fig:mainframework}.

\subsubsection{Cross-Component Attention.}
To better capture dependencies and synergy between different creative components, we introduce cross-component attention. For each creative component $C_j$ and its k-th element candidate $e_k^{i,j}$, we compute a cross-component context vector $\mathbf{z}_{k}^{i,j}$:
\begin{equation}
\mathbf{z}_{k}^{i,j} = \mathrm{Attn}(\text{emb}(e_k^{i,j}), \{ \text{emb}(e_k^{i,j}) \} \cup \bigcup_{j' \neq j} \{\text{emb}(e_l^{i,j'})\}_{l=1}^{n_{i,j'}}),
\end{equation}
$\mathrm{Attn}$ denotes a modified multi-head attention mechanism~\cite{Vaswani2017AttentionIA}.
Here, $\text{emb}(e_k^{i,j})$, the embedding of k-th element of creative component $C_j$, acts as the query. Its own embedding and the embeddings of all candidate elements from other components $\bigcup_{j' \neq j} \{\mathrm{emb}(e_l^{i,j'})\}_{l=1}^{n_{i,j'}}$ form the keys and values.
This enables the selection probability of each $e^{i,j}_k$ to be explicitly conditioned on the global candidate context, facilitating the modeling of inter-element interactions.

\subsubsection{Non-Autoregressive Context-Aware Generation.}
Given an advertisement $A_i$, candidate pools for each creative component $\{S_{i,j}\}_{j=1}^m$, and contextual features $q$ (including user, query, and environment signals), we seek to generate a creative combination tuple $\mathbf{c}_i = (c_{i,1}, ..., c_{i,m})$,
where $c_{i,j} \in S_{i,j}$. To ensure efficiency and scalability, we employ a non-autoregressive generation strategy: all elements are selected in parallel, conditioned on a shared context embedding $\mathbf{h}_i$. The generative probability is:
\begin{align}
P(\mathbf{c}_i \mid A_i, \{S_{i,j}\}_{j=1}^m, q) = \prod_{j=1}^m P(c_{i,j} \mid \mathbf{h}_i, S_{i,j}),
\end{align}
where $\mathbf{h}_i = \text{Enc}(A_i, q, \{\text{stat}(S_{i,j})\})$ is a contextual embedding produced by an encoder network (e.g., Transformer or MLP) that integrates information from the ad, request, and candidate pool.

For each creative component $C_j$, the selection probability for k-th candidate $e_k^{i,j}$ is computed as:
\begin{align}\label{eq:normalize_2}
& P(c_{i,j} = e_k^{i,j} \mid \mathbf{h}_i, S_{i,j}) \notag \\
&= 
\frac{
    \exp\left( \phi\left(\mathbf{h}_i,\, \text{emb}(e_k^{i,j}),\, \text{emb}(C_j),\mathbf{z}_{k}^{i,j}\right) \right)
}{
    \sum\limits_{l=1}^{n_{i,j}} 
    \exp\left( \phi\left(\mathbf{h}_i,\, \text{emb}(e_l^{i,j}),\, \text{emb}(C_j),\mathbf{z}_{k}^{i,j}\right) \right)
}.
\end{align}
where $\phi$ denotes a shared context-aware generation function for all creative components, and $\text{emb}(e_k^{i,j})$, $\text{emb}(C_j)$ are learnable embeddings of the k-th element and its type, respectively. This mechanism aligns with the context-sensitive sampling in language models, allowing top-$K$ combinations to be sampled efficiently according to their probabilities.

\textbf{Multi-Instance Reward Learning.}
To address the challenge of sparse, bag-level feedback (i.e., click signals observed only at the combination level), we adopt a multi-instance learning (MIL) framework. Here, each sampled creative combination $\mathbf{c}_i$ is viewed as a ``bag'', while its constituent elements $\{c_{i,j}\}_{j=1}^m$ are regarded as ``instances''. The feedback signal (e.g., click or non-click) for the bag is attributed to individual instances via a shared policy network, ensuring that both element- and combination-level representations reside in a unified feature space.

Let $\mathbf{f}_{e_k^{i,j}}$ denote the feature representation of element $e_k^{i,j}$, and $\mathbf{f}_{\mathbf{c}_i}$ that of the combination. They are defined as:
\begin{align}
\mathbf{f}_{e_k^{i,j}} = \left(\text{emb}(e_k^{i,j}),\, \text{emb}(C_j),\mathbf{z}_{k}^{i,j}\right), \\
\mathbf{f}_{\mathbf{c}_i} = \sum_{j=1}^m w_{i,j}( \mathbf{h}_i, \bigcup^{m}_{j=1}\{\mathbf{f}_{e_{l}^{i,j}}\}_{l=1}^{n_{i,j}} ) \cdot \mathbf{f}_{e_{k}^{i,j}}.
\end{align}
where $w_{i, j}$ is a learnable context-dependent attention weight for each creative element, and is set zero for unexposed elements.
As illustrated in Figure \ref{fig:mainframework}, the shared policy network $\pi_\theta$ acts as the shared context-aware generation function $\phi$, which predicts scores for both levels:
\begin{align}
s_m^i= \pi_\theta(\mathbf{h}_i, \mathbf{f}_{\mathbf{c}_i}), \\
p_k^{i,j} = \pi_\theta(\mathbf{h}_i, \mathbf{f}_{e_k^{i,j}}).
\end{align}
The MIL loss integrates both bag- and instance-level objectives:
\begin{align}
\mathcal{L}_{\mathrm{MIL}} = \mathcal{L}_{\mathrm{comb}} + \mathcal{L}_{\mathrm{ele}}
\end{align}
where $\mathcal{L}_{\mathrm{comb}}$ and $\mathcal{L}_{\mathrm{ele}}$ are cross-entropy losses for combinations and elements, respectively. 
This joint training approach allows the element-level model to benefit from combination-level (bag) feedback, while the combination-level model also leverages fine-grained predictions from individual elements.
In this way, both levels reinforce each other, leading to better learning and generalization.

\subsubsection{Reinforcement Learning for Sparse Supervision.}
Given the extreme sparsity of creative combination data, the need to propagate rewards to unexposed elements and combinations,
and the challenge of disentangling creative effectiveness from product appeal,
using traditional supervised learning paradigms leads to serious training performance degradation.
We construct the generative selection process using policy gradient methods. Here, the policy network $\pi_\theta$ outputs a categorical distribution for each component, utilizing observed or predicted rewards (e.g., observed clicks or predicted click-through rates) for credit assignment.

The policy network $\pi_\theta$ generates combinations by sampling elements for each component from their respective distributions:
\begin{align}
\pi_\theta(\mathbf{c}_i \mid A_i, \{S_{i,j}\}_{j=1}^m, q) = \prod_{j=1}^m \pi_\theta(c_{i,j} \mid A_i, S_{i,j}, q)
\end{align}
The policy gradient objective is:
\begin{align}
& \nabla_\theta J(\pi_\theta) \notag =\\
& 
\mathbb{E}_{\mathbf{c}_i \sim \pi_\theta} \left[ r(A_i, \mathbf{c}_i, q) \nabla_\theta \log \pi_\theta(\mathbf{c}_i \mid A_i, \{S_{i,j}\}_{j=1}^m, q) \right]
\end{align}
where $r(A_i, \mathbf{c}_i, q)$ denotes the observed or predicted reward for the combination. This policy gradient mechanism allows the model to learn from sparse feedback by attributing credit to the sampled combinations based on their rewards.
This precise attribution is crucial for disentangling creative effectiveness from overall product appeal,
thereby enhancing the ability to generalize to rarely or never exposed elements and combinations.

This reinforcement learning approach enables the model to dynamically adjust its generation policy based on observed rewards, thereby substantially mitigating cold-start problems and improving sample efficiency. Through leveraging the policy gradient method, the model can effectively learn to generate high-quality creative combinations under sparse supervision.

\begin{table*}[t]
\centering
\resizebox{.7\linewidth}{!}{
\begin{tabular}{c|cccccc}
\toprule
                  & title       & image       & marketing  & highlight   & attribute  & structure\_attribute \\ \midrule
\# ids            & 230 million & 140 million & 15 million & 300 million & 70 million & 320 million          \\
coverage          & 100.0\%      & 100.0\%      & 39\%       & 95.6\%      & 58.3\%     & 67.8\%               \\
\# avg\_candidate & 6.3         & 3.4         & 2.2        & 7.0         & 2.3        & 8.1                  \\
\# expo\_per\_day  & 2.3         & 3.6         & 11.5       & 1.9         & 1.3        & 1.0                  \\ \bottomrule
\end{tabular}
}
\caption{
Training dataset statistics.
``\# ids'': number of unique IDs, ``coverage'' :\% of ads with the component, ``\# avg\_candidate'': average candidates per ad, ``\# expo\_per\_day'': average daily exposures per candidate.
}
\label{tab:creative_element}
\end{table*}

\textbf{Training and Inference.}
The overall training objective is a weighted sum of the MIL and RL losses:
\begin{align}
\mathcal{L} = \mathcal{L}_{\mathrm{MIL}} + \lambda \mathcal{L}_{\mathrm{RL}}
\end{align}
where $\lambda$ balances supervised and reinforcement learning signals. During inference and online serving, the generative model samples $K$ diverse candidate combinations in parallel (via the context-aware sampling distribution), each evaluated by a reward/CTR model, and selects the highest-scoring combination:
\begin{align}
\{\mathbf{c}_i^{(k)}\}_{k=1}^K \sim \pi_\theta(\cdot \mid A_i, \{S_{i,j}\}_{j=1}^m, q), \\
\mathbf{c}_i^* = \arg\max_{k} \mathrm{CTR}(A_i, \mathbf{c}_i^{(k)}, q).
\end{align}

Overall, our paradigm enables efficient modeling and generation of contextually optimal creative combinations, while substantially enhancing the robustness and generalization of creative selection through dense, context-aware, and reward-driven training signals.

\begin{table}[t]
\centering
\resizebox{.5\linewidth}{!}{
\begin{tabular}{l|cc}
\toprule
& CTR & RPM \\ \midrule
Uniform & - & - \\
+DCO & +0.76\%  & +0.26\%   \\
+MLP & +1.25\%  & +0.53\%   \\ \midrule
+$f_{\mathrm{ele}}$    & +2.47\%     &  +1.91\%         \\ 
+$f_{\mathrm{comb}}$    & +0.96\%     & +1.18\%         \\
+GenCO    & +3.88\%       & +3.26\%           \\ \bottomrule\end{tabular}
}
\caption{Online performance of different deployed methods. Uniform uses a random policy which applies no creative selection at all, $f_{\mathrm{ele}}$ refers to  context-aware   generation function, and $f_{\mathrm{comb}}$ refers to combination-level reward function.}
\label{tab:online_performance}
\end{table}

\section{Experiments}
\label{sec:exp}

\subsection{Experimental Setup}

\subsubsection{Dataset.}

For offline evaluation, we constructed our dataset by randomly sampling from online system logs. The training set, collected between June 20th and July 22nd, contains data from 275 million users and 110 million ads, with an average CTR of 6.84\%. Detailed statistics are provided in Table~\ref{tab:creative_element}. 
To ensure unbiased offline testing, we collected a separate test dataset from July 23rd to July 24th, covering 6.2 million users and 7.2 million ads. Importantly, the offline test set originates from 5\% of online traffic, collected under a random policy, which ensured each creative element had an equal probability of exposure for any given ad. This randomized exposure policy enables fair and accurate estimation of CTR for each candidate, thereby improving the reliability of our offline evaluation and reducing the estimation variance compared to Inverse Propensity Weighting methods~\citep{glynn2010introduction}.
For online A/B testing, model training is performed daily, utilizing data from the most recent 33 days.
See the Appendix for the implementation details.

\begin{table*}[t]
\centering
\resizebox{.65\linewidth}{!}{
\begin{tabular}{@{}c|cccccc@{}}
\toprule
& title    & image    & marketing & highlight & attribute & structure\_attribute \\ \midrule
w/o MIL   & -0.126\% & -0.084\% & -0.165\%  & 0.071\%   & -0.010\%   & -0.128\%  \\
w/o RL  & -0.553\% & -1.882\% & -0.025\%  & -0.106\%  & -0.088\%   & -0.378\%    \\
w/o context  & -0.044\%  & -0.070\% & -0.283\%  & 0.234\%   & -0.044\%   & -0.158\%     \\
w/o all  & -0.640\% & -1.914\% & -0.129\%  & 0.074\%   & -0.026\%   & -0.041\%   \\ \bottomrule
\end{tabular}
}
\caption{
Ablation study on the key modules of GenCO. It shows the performance impact of removing Multi-Instance Learning (w/o MIL), Reinforcement Learning (w/o RL), Cross-Component Attention (w/o context), and all three components (w/o all).
}
\label{tab:ablation_study}
\vspace{-2mm}
\end{table*}

\begin{table*}[t]
\centering
\resizebox{.65\linewidth}{!}{
\begin{tabular}{@{}ccccccc@{}}
\toprule
\# candidate                 & title   & image   & marketing & highlight & attribute & structure\_attribute \\ \midrule
\textless{}3 (\textless{}4)       & +1.23\% & +1.15\% & +2.36\%   & +0.16\%   & +2.11\%   & +2.43\%              \\
3-5 (4-8)                         & +4.00\% & +6.40\% & +3.15\%   & +1.37\%   & +2.78\%   & +2.07\%              \\
\textgreater{}5 (\textgreater{}8) & +2.76\% & +8.34\% & +3.02\%   & +3.09\%   & +3.00\%   & +5.10\%              \\ \bottomrule
\end{tabular}
}
\vspace{-2mm}
\caption{CTR improvement relative to the random policy, segmented by the number of candidates per creative component. The candidate pools are grouped into three brackets: [\textless3, 3-5, \textgreater5] for title, image, and marketing; and [\textless4, 4-8, \textgreater8] for the others.
}
\label{tab:ctr_lift_by_candi_num}
\end{table*}

\subsubsection{Baselines.}
To comprehensively evaluate our approach, we compare it against several representative algorithms, with all parameters carefully tuned to maximize overall  CTR.

\textbf{Uniform}: This baseline randomly selects creatives for each advertisement, without utilizing any prior knowledge or historical data. As a result, it consistently exhibits suboptimal performance on test sets and serves as a lower bound for comparison.

\textbf{MLP}~\citep{rosenblatt1958perceptron}: This method uses a Multi-Layer Perceptron to predict the CTR of each possible creative combination. The combination with the highest predicted CTR is selected for exposure, representing a standard supervised learning approach in creative selection.

\textbf{DCO}~\citep{wei2022towards}: Dynamic Creative Optimization adopts a post-auction strategy, employing successive elimination to rank and select creative combinations based on their observed CTRs. This method reflects a more advanced optimization paradigm for creative selection.

\begin{figure}[t]
    \centering
    \includegraphics[width=1\linewidth]{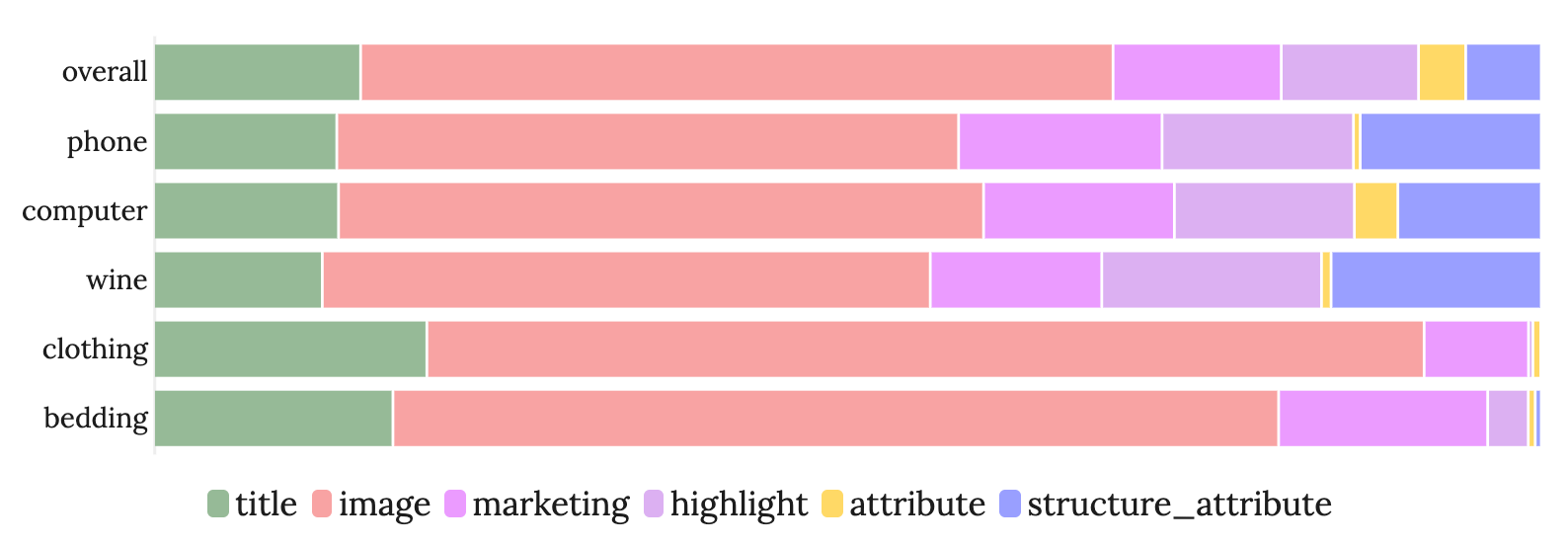}
    \caption{
    Relative importance of creative components across different product categories. Importance is determined by the model's learned attention weights.
    }
    \label{fig:importance}
    \vspace{-5mm}
\end{figure}

\subsubsection{Evaluation metrics.}

For offline evaluation, we utilize Simulated CTR (sCTR)~\cite{wang2021hybrid}, a metric that accumulates impressions and clicks when generated creatives match those exposed online. To ensure bias avoidance and result comparability, we propose the calibrated sCTR. This enhancement involves (1) calibrating sCTR based on the number of candidate creative elements, and (2) computing it on an offline test dataset generated using a random policy. This calibrated sCTR is then applied to multiple creative component evaluations, where we compute component-level sCTR by assessing one component at a time.

For calculating calibrated sCTR, we use an offline test dataset collected via a random policy, specifically focusing on creative component $C_j$. The dataset provides:
(1) An ad set $\mathbf{A} = \{A_r\}_{r=1}^{R}$ and a candidate element set $S_{r, j} = \{ e^{r,j}_{1},  e^{r,j}_{2}, \dots,  e^{r,j}_{n_{r,j}} \}_{r=1}^{R}$, where each ad $A_r$ has a real creative component exposure $c_{r,j}$;
(2) Ad click labels $\mathbf{Y} = \{y_r\}_{r=1}^{R}$;
(3) A creative selection model $\mathbf{f}$;
(4) A selection policy $\pi_\theta$.
Calibrated sCTR is computed by initializing exposure and click to zero. For each ad $A_r$: we predict element preference scores for each candidate element $e_k^{r,j}$ using $\mathbf{f}$ to obtain $p_k^{r,j}$, then select a creative element $e_{\hat{k}}^{r,j}$ according to $\pi_\theta$. If $e_{\hat{k}}^{r,j}$ matches the exposed element $c_{r,j}$, we increment exposure by $n_{r,j}$ and click by $n_{r,j} \times y_r$. Finally, calibrated sCTR is the ratio of total clicks to total exposures.
The detailed procedure is outlined in Algorithm~\ref{alg:sCTR}.

\begin{algorithm}[t]
\caption{Evaluation Metric - calibrated sCTR}
\label{alg:sCTR}
\begin{algorithmic}[1]
\State Initialize $\mathrm{exposure} \gets 0$, $\mathrm{click} \gets 0$
\For{$r \gets 1$ to $R$}
    \State Retrieve candidate element set $S_{r,j}$ for ad $A_r$
    \For{$k \gets 1$ to $n_{r,j}$}
        \State Predict element preference score $p_k^{r,j} \gets \mathbf{f}(e_k^{r,j})$
    \EndFor
    \State Select creative $e_{\hat{k}}^{r,j} \gets \pi_\theta(\{p_k^{r,j}\}_{k=1}^{n_{r,j}})$
    \If{$e_{\hat{k}}^{r,j} = c_{r,j}$}
        \State $\mathrm{exposure} \gets \mathrm{exposure} + n_{r,j}$
        \State $\mathrm{click} \gets \mathrm{click} + n_{r,j} \times y_r$
    \EndIf
\EndFor
\State \Return sCTR $= \mathrm{click} / \mathrm{exposure}$
\end{algorithmic}
\end{algorithm}

\subsubsection{Implementation Details.}
The input features for GenCO consist of four parts:
(1) user features: user hash ID, age, gender;
(2) ad features: ad hash ID;
(3) query features: query hash ID and query text segments;
(4) creative element features: elements from six different creative components --- title, image, marketing, highlight, attribute, and structure\_attribute --- are each represented by a hashed ID.
Ad, query, and creative element IDs are each mapped to a separate hash embedding with a dimension of 4. The sparse hash size for each component is determined by the number of its distinct ids present in the training data.

The number of candidates for each creative component is padded or truncated to a fixed size of 15. The model's architecture is defined as follows. The context encoder network is a 3-layer MLP with dimensions [32, 16, 8], which applies Batch Normalization and produces a contextual embedding $\mathbf{h}_i$ of dimension 8. The cross-component attention module is implemented using masked attention with a hidden dimension of 4. The shared context-aware generation function $\phi$ is a 3-layer MLP with dimensions [64, 32, 8] and a final output layer of dimension 1. All hidden layers across the model use the ReLU activation function.

The training data is arranged in chronological order. To enable the model to adapt to constantly changing creative candidates and prioritize recent performance, we apply the SGD optimizer to all model parameters. The training hyperparameters include a batch size of 1024 and a fixed learning rate of 5e-3, with no learning rate scheduler. For policy gradient optimization, clicked creative combinations are assigned a reward of 1, while non-clicked creative combinations receive a reward of -0.1. The hyperparameter $\lambda$ used to balance the MIL and RL losses is set to 0.15.

In our online experiments, we evaluate user engagement with Click-Through Rate (CTR) and platform revenue with Revenue Per Mile (RPM). For offline evaluation, we observed a discrepancy between traditional metrics like AUC and recall and our online results. This led us to adopt calibrated sCTR, a metric specifically chosen to better align with online creative performance.

\subsection{Offline Evaluation}

\subsubsection{Ablation Study.}
The key of our model design lies in three parts: multi-instance learning is to address feedback signal ambiguity, reinforcement learning aims to alleviate label sparsity and promote creative diversity, and cross-component attention modeling to take global candidate context and inter-element interactions into consideration. 
To validate the effectiveness of each of them, we perform an ablation analysis, and results are shown in Table~\ref{tab:ablation_study}.
All these models are trained using the same training policy including optimizer, learning rate and etc.
The results indicate that removing any of these components generally leads to a performance decline, confirming the importance of each component in our model. Notably, the absence of reinforcement learning (w/o RL) and the absence of all components (w/o all) result in the most significant negative impact on CTR, highlighting their critical roles in the model's effectiveness.

\subsubsection{Importance of components.}

To ascertain the relative importance of each creative component, we analyze the learned context-dependent attention weight $w_{i, j}$ from multi-instance learning, used for aggregating different exposed components, as illustrated in Figure~\ref{fig:importance}. Our findings indicate that the visual image component consistently holds the highest importance
across all categories.
Further analysis of the attention weights in categories such as the phone, computer, wine, clothing, and bedding reveals distinct user preferences: users tend to focus more on the highlight, attribute, and structure\_attribute components in categories like the phone, computer, and wine, whereas in the categories of clothing and bedding, greater emphasis is placed on ad images. These results underscore the overarching significance of the image component and suggest that increasing the diversity of image candidates --- currently averaging only 3.4 per ad --- could unlock further performance gains.

\subsubsection{CTR improvement vs candidate number.}
We analyze the impact of creative selection across different candidate number buckets. The CTR lift in each bucket for every creative component is shown in Table~\ref{tab:ctr_lift_by_candi_num}. The degree of improvement increases with the number of candidates, demonstrating the advantages of our generative model and two-stage approach.
Our generative model excels with a larger number of candidates, significantly outperforming the baseline. This is because the generative paradigm has a higher performance ceiling compared to the discriminative paradigm.
The two-stage framework, involving generative modeling followed by reward modeling, effectively balances efficiency and effectiveness.
For instance, the image component shows the most significant improvement, with CTR lifts increasing as the candidate pool grows. Other components like titles, marketing words, highlights, attributes, and structure attributes also exhibit consistent improvements across different candidate number buckets. This underscores the model's robustness in handling various creative components and its ability to generate and select optimal candidates from larger pools.

\subsection{Online Evaluation}
The online A/B test results, summarized in Table~\ref{tab:online_performance}, confirm the significant real-world impact and commercial value of our GenCO framework when deployed in a live industrial environment.
While standard MLP and DCO methods provide modest, incremental gains over the Uniform random baseline, our approach demonstrates a clear performance hierarchy.
The introduction of the context-aware generation function ($f_{\mathrm{ele}}$) alone yields a substantial uplift in both CTR and RPM (Revenue Per Mille), with the combination-level reward function ($f_{\mathrm{comb}}$) also providing a significant boost.
Ultimately, the fully-integrated GenCO framework achieves the highest performance, delivering a remarkable relative lift of +3.88\% in CTR and +3.26\% in RPM.
GenCO has now been successfully serving the main traffic of the large-scale online system, which translates directly into significant financial gains for the leading e-commerce platform.

\section{Conclusion}
In this paper, we presented GenCO, a novel generative framework that integrates generative modeling with multi-instance reward learning for creative optimization in e-commerce advertising. By modeling creative selection as a context-aware generation process and leveraging reinforcement learning and multi-instance learning, our approach effectively addresses the challenges of combinatorial explosion, sparse user feedback, and disentangling creative impact from product appeal.  Our architecture efficiently generates diverse and coherent creative combinations and accurately attributes rewards to individual elements, enabling robust optimization under real-world, large-scale constraints. Through extensive offline experiments and online deployment on a leading e-commerce platform, GenCO demonstrated significant improvements in both click-through rate (CTR) and advertising revenue compared to existing baselines. Our ablation studies further confirmed the critical roles of multi-instance learning, reinforcement learning, and cross-component attention in achieving these results. Overall, our work provides a scalable and effective solution for creative optimization, paving the way for more intelligent and data-driven advertising strategies in e-commerce.

\bibliography{aaai2026}

\end{document}